\documentclass{article}

\usepackage[longnamesfirst, authoryear, square, compress]{natbib}

\usepackage[final]{neurips_2018}

\usepackage[utf8]{inputenc} 
\usepackage[T1]{fontenc}    
\usepackage{url}            
\usepackage{booktabs}       
\usepackage{amsfonts}       
\usepackage{nicefrac}       
\usepackage{microtype}      
\usepackage{mathrsfs}
\usepackage{hhline}
\usepackage{bm}
\usepackage{bbm} 
\usepackage{tabularx}
\usepackage{multirow}
\usepackage{graphicx} 
\usepackage{amsmath,amsfonts,amssymb}
\usepackage{wrapfig}

\title{MacNet: Transferring Knowledge from Machine Comprehension to Sequence-to-Sequence Models}

\author{Boyuan Pan$^\dag$, Yazheng Yang$^\ddag$, Hao Li$^\dag$, Zhou Zhao$^\ddag$, Yueting Zhuang$^\ddag$, Deng Cai$^{\dag \sharp}$\thanks{corresponding author}, Xiaofei He$^{\star \dag}$\\
	$^\dag$State Key Lab of CAD$\&$CG, Zhejiang University\\
	$^\ddag$College of Computer Science, Zhejiang University\\
	$^\sharp$Alibaba-Zhejiang University Joint Institute of Frontier Technologies\\
	$^\star$Fabu Inc., Hangzhou, China\\
	{\tt $\{$panby, yazheng\_yang, haolics, zhaozhou, yzhuang, dcai$\}$@zju.edu.cn}\\
	{\tt xiaofeihe@fabu.ai} \\
}

\begin{document}
	
	\maketitle

\begin{abstract}
	Machine Comprehension (MC) is one of the core problems in natural language processing, requiring both understanding of the natural language and knowledge about the world. Rapid progress has been made since the release of several benchmark datasets, and recently the state-of-the-art models even surpass human performance on the well-known SQuAD evaluation. In this paper, we transfer knowledge learned from machine comprehension to the sequence-to-sequence tasks to deepen the understanding of the text. We propose \emph{MacNet}: a novel encoder-decoder supplementary architecture to the widely used attention-based sequence-to-sequence models. Experiments on neural machine translation (NMT) and abstractive text summarization show that our proposed framework can significantly improve the performance of the baseline models, and our method for the abstractive text summarization achieves the state-of-the-art results on the \emph{Gigaword} dataset.
\end{abstract}

\section{Introduction}

Machine comprehension (MC) has gained significant popularity over the past few years and it is a coveted goal in the field of natural language understanding. Its task is to teach the machine to understand the content of a given passage and then answer a related question, which requires deep comprehension and accurate information extraction towards the text. With the release of several high-quality benchmark datasets~\citep{hermann2015teaching,rajpurkar2016squad,joshi2017triviaqa}, end-to-end neural networks~\citep{rnet,xiong2017dynamic,cui2017attention} have achieved promising results on the MC tasks and some even outperform humans on the SQuAD~\citep{rajpurkar2016squad}, which is one of the most popular machine comprehension tests. Table \ref{tab1} shows a simple example from the SQuAD dataset.

Sequence-to-sequence (seq2seq) models~\citep{sutskever2014sequence} with attention mechanism~\citep{bahdanau2015neural}, in which an encoder compresses
the source text and a decoder with an attention mechanism generates target words, have shown great capability to handle many natural language generation tasks such as machine translation~\citep{luong2015effective,xia2017deliberation}, text summarization~\citep{rush2015neural,nallapati2016abstractive} and dialogue systems~\citep{williams2017hybrid}, \emph{etc}. However, these encoder-decoder networks directly map the source input to a fixed target sentence to learn the relationship between the natural language texts, which makes them hard to capture
a lot of deep intrinsic details and understand the potential implication of them~\citep{li2017modeling,shi2016does}.

Inspired by the recent success of the approaches for the machine comprehension tasks, we focus on exploring whether MC knowledge can further help the attention-based seq2seq models deeply comprehend the text. Machine comprehension requires to encode words from the passage and the question firstly, then many methods~\citep{seo2017bidirectional,wang2016multi,xiong2018dcn} employ attention mechanism with an RNN-based modeling layer to capture the interaction among the passage words conditioned on the question and finally use an MLP classifier or pointer networks~\citep{vinyals2015pointer} to predict the answer span. The MC-encoder mentioned above is a common component in the seq2seq models, while the RNN-based modeling layer whose input is the attention vectors is also supposed to augment the performance of the outputs of the seq2seq models. Intuitively, MC knowledge could improve seq2seq models through measuring the relevance between the generated sentence and the input source. Moreover, while quesiton answering and text generation have different training data distributions, they can still benefit from sharing their model's high-level semantic components~\citep{guo2018soft}.

In this paper, we propose \emph{MacNet}, a machine comprehension augmented encoder-decoder supplementary architecture that can be applied to a variety of sequence generation tasks. We begin by pre-training an MC model that contains both the RNN-based encoding layer and modeling layer as the transferring source. In the sequence-to-sequence model, for encoding, we concatenate the outputs of the original encoder and the transferred MC encoder; for decoding, we first input the attentional vectors from the seq2seq model into the transferred MC modeling layer, and then combine its outputs with the attentional vectors to formulate the predictive vectors. Moreover, to solve the class imbalance resulted by the high-frequency phrases, we adopt the \emph{focal loss}~\citep{lin2017focal} which reshapes the standard cross entropy to improve the weights of the loss distribution.

To verify the effectiveness of our approach, we conduct experiments on two representative sequence generation tasks.

(1) \emph{Neural Machine Translation}. We transfer the knowledge from the machine comprehension model to the attention-based Neural Machine Translation (NMT) model. Experimental results show that our method significantly improves the performance on several large-scale MT datasets.

(2) \emph{Abstractive Text Summarization}. We modify the Pointer-Generator Networks recently proposed by \citet{see2017get}. We evaluate this model on the \emph{CNN/Daily Mail}~\citep{hermann2015teaching} and \emph{Gigaword}~\citep{rush2015neural} datasets. Our model obtains 37.97 ROUGE-1, 18.16 ROUGE-2 and 34.93 ROUGE-L scores on the English \emph{Gigaword} dataset, which is an improvement over previous state-of-the-art results in the literature.

\begin{table*}
	\begin{center}
		\begin{tabular}{l}
			\toprule
			\textbf{Passage}: This was the first Super Bowl to feature a quarterback on both teams who was the \#1\\ pick in their draft classes. \textbf{Manning was the \#1 selection of the 1998 NFL draft}, while Newton\\ was picked first in 2011. The matchup also pits the top two picks of the 2011 draft against each\\ \vspace{1mm} other: Newton for Carolina and Von Miller for Denver.\\ \vspace{1mm}
			{\textbf{Question}: Who was considered to be the first choice in the NFL draft of 1998?}\\
			{\textbf{Answer}: Manning}\\ 
			\bottomrule
		\end{tabular}
	\end{center}
	\caption{\label{tab1} An example from the SQuAD dataset.}       
\end{table*}

\section{Related Work}
\subsection{Machine Comprehension}
Teaching machines to read, process and comprehend text and then answer questions, which is called machine comprehension, is one of the key problems in artificial intelligence. Recently, \citet{rajpurkar2016squad} released the Stanford Question Answering Dataset (SQuAD), which is a high-quality and large-scale benchmark, thus inspired many significant works~\citep{xiong2017dynamic,pan2017memen,cui2017attention,seo2017bidirectional,wang2016multi,xiong2018dcn,shen2017reasonet,rnet}. Most of the state-of-the-art works are attention-based neural network models. \citet{seo2017bidirectional} propose a bi-directional attention flow to achieve a query-aware context representation. \citet{rnet} employ gated self-matching attention to obtain the relation between the question and passage, and their model is the first one to surpass the human performance on the SQuAD. In this paper, we show that the pre-trained MC architecture can be transferred well to other NLP tasks.

\subsection{Sequence-to-sequence Model}
Existing sequence-to-sequence models with attention have focused on generating the target sequence by aligning each generated output token to another token in the input sequence. This approach has proven successful in many NLP tasks, such as neural machine translation~\citep{bahdanau2015neural}, text summarization~\citep{rush2015neural} and dialogue systems~\citep{williams2017hybrid}, and has also been adapted to other applications, including speech recognition~\citep{chan2016listen} and image caption generation~\citep{xu2015show}. In general, these models encode the input sequence as a set of vector representations using a recurrent neural network (RNN). A second RNN then decodes the output sequence step-by-step, conditioned on the encodings. In this work, we augment the natural language understanding of this encoder-decoder framework via transferring knowledge from another supervised task.

\subsection{Transfer Learning in NLP}
Transfer learning, which aims to build learning
machines that generalize across different domains following
different probability distributions, has been widely applied in natural language processing tasks~\citep{collobert2011natural,glorot2011domain,min2017question,mccann2017learned,pan2018discourse}. \citet{collobert2011natural} propose a unified neural network architecture and learned from unsupervised learning that can be applied to various
natural language processing tasks including part-of-speech tagging, chunking, named entity recognition, and semantic role labelling.  \citet{glorot2011domain} propose a deep learning approach which learns to extract a meaningful representation for each review in an unsupervised fashion. \citet{mccann2017learned} propose to transfer the pre-trained encoder from the neural machine translation (NMT) to the text classification and question answering tasks. \citet{pan2018discourse} propose to transfer the encoder of a pre-trained discourse marker prediction model to the natural language inference model. Unlike previous works that only focus on the encoding part or unsupervised knowledge source, we extract multiple layers of the neural networks from the machine comprehension model and insert them into the sequence-to-sequence model. Our approach not only makes the transfer more directly compatible with subsequent RNNs, but also augments the text understanding of the attention mechanism.

\section{Machine Comprehension Model}
\subsection{Task Description}
In the machine comprehension task, we are given a question $\mathbf{Q} = \{ \mathbf{q}_1, \mathbf{q}_2, ..., \mathbf{q}_m\}$ and a passage $\mathbf{P} = \{\mathbf{p}_1, \mathbf{p}_2, ..., \mathbf{p}_n \}$, where $m$ and $n$ are the length of the question and the passage. The goal is to predict the correct answer $\mathbf{a}^c$ which is a subspan of $\mathbf{P}$.

\subsection{Framework}
\label{s2.2}
The state-of-the-art MC models are various in structures, but many popular works are essentially the combination of the encoding layer, the attention mechanism with and an RNN-based modeling layer and the output layer\citep{wang2016multi,seo2017bidirectional,pan2017memen,xiong2018dcn}. Now we describe our MC model as follows.

\textbf{Encoding Layer } We use pre-trained word vectors \emph{GloVe}~\citep{pennington2014glove} and character-level embeddings to transfer the words into vectors, where the latter one applies CNN over the characters of each word and is proved to be helpful in handling out-of-vocab words~\citep{kim2014convolutional}. We then use a bi-directional LSTM on top of the concatenation of them to model the temporal interactions between words:
\begin{equation}
\label{eq1}
\begin{aligned}
&\mathbf{u}_i = G_{enc} (f_{rep}(\mathbf{q}_i),\mathbf{u}_{i-1}), i= 1,...,m \\
&\mathbf{h}_j = G_{enc} (f_{rep}(\mathbf{p}_j),\mathbf{h}_{j-1}), j= 1,...,n
\end{aligned}
\end{equation}
where $G_{enc}$ is the bi-directional LSTM, $f_{rep}(\mathbf{x}) =[{\rm Glove}(\mathbf{x}); {\rm Char}(\mathbf{x})]$ is the concatenation of the word and character embedding vectors of the word $\mathbf{x}$; $\{ \mathbf{u}_i \}_{i=1}^{m}$ and $\{ \mathbf{h}_j \}_{j=1}^{n}$ are the contextual representations of the question $\mathbf{Q}$ and the passage $\mathbf{P}$.\\

\textbf{Attention Layer } Attention mechanisms are commonly used in machine comprehenion to model the document so that its representation can emphasize the key information and capture long-distance dependencies: 
\begin{equation}
\label{eq2}
\begin{aligned}
\mathbf{G} = f_{att}(\{\mathbf{u}_i\}_{i=1}^{m}, \{\mathbf{h}_j\}_{j=1}^{n})
\end{aligned}
\end{equation}
Here, the attention function $f_{att}$ represents a series of normalized linear and logical operations. We follow \citep{seo2017bidirectional} to use a bi-directional attention flow (BiDAF), where the passage and the question  are interacted each other with an alignment matrix, $\mathbf{G}$ is the query-aware context representation.\\

\textbf{Modeling Layer } In this step, we use the stacking LSTM on $\mathbf{G}$ to further capture the interaction among the passage words conditioned on the question:
\begin{equation}
\label{eq3}
\begin{aligned}
\mathbf{m}_j = G_{model} (\mathbf{G}_j,\mathbf{m}_{j-1}), j= 1,...,n 
\end{aligned}
\end{equation}
where $G_{model}$ is two layers of uni-directional LSTM, each $\mathbf{m}_j$ is expected to represent the contexual information of the $j$-th word in the passage to the whole question.

We use a simple MLP classifier on the combination of  $\{\mathbf{m}_j\}_{j=1}^{n}$ and $\mathbf{G}$ to locate the start and end positions of the answer. For training, we define the training loss as the sum of the negative log probability of the true positions by the predicted distributions.

\begin{figure*}[t]
	\begin{center}
		\includegraphics[height=0.41 \textwidth]{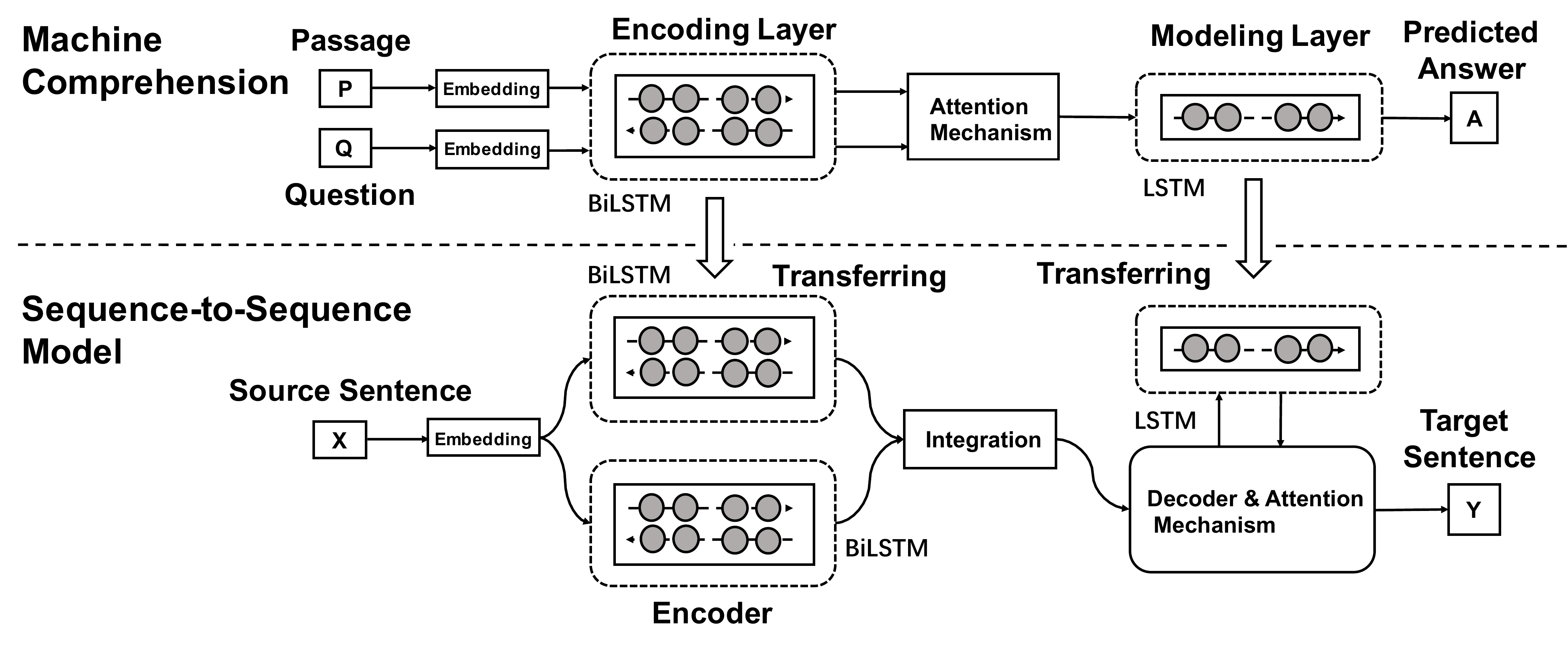}
		\caption{\label{fig1} Overview of our MacNet framework, comprising the part of Machine Comprehension (upper) for pre-training and Sequence-to-Sequence model (bottom) to which the learned knowledge will be transferred.}
	\end{center}
\end{figure*}

\section{MacNet Architecture}
In this section, as shown in the Figure \ref{fig1}, we introduce how our MacNet transfers the knowledge from the MC model to the seq2seq model. The sequence-to-sequence models are typically implemented with a Recurrent Neural Network (RNN)-based encoder-decoder framework. Such a framework directly models the probability $P(y|x)$ of a target sentence $y = \{y_1, y_2, ..., y_{T_y}\}$ conditioned on the source sentence $x = \{x_1, x_2, ..., x_{T_x}\}$, where $T_x$ and $T_y$ are the length of the sentence $x$ and $y$. 

\subsection{Encoder}
For the seq2seq model, the encoder reads the source sentence $x$ word by word and generates a hidden representation of each word $x_s$: 
\begin{equation}
\label{eq4}
\begin{aligned}
\tilde{\mathbf{h}}_s = F_{enc}({\rm Emb}(x_s),\tilde{\mathbf{h}}_{s-1})
\end{aligned}
\end{equation}
where $F_{enc}$ is the recurrent unit such as Long Short-Term Memory (LSTM)~\citep{sutskever2014sequence} unit or Gated Recurrent Unit (GRU)~\citep{cho2014learning}, ${\rm Emb}(x_s)$ is the embedding vector of $x_s$, $\tilde{\mathbf{h}}_s$ is the hidden state. In this paper, we use the bi-directional LSTM as the recurrent unit to be consistent with the encoding layer of the MC model described in Section \ref{s2.2}. 

To augment the performance of the encoding part, we use a simple method to exploit the word representations that learned from the MC task. For the source sentence $x$, we use the bi-directional LSTM of the equation (\ref{eq1}) as another encoder and obtain:
\begin{equation}
\begin{aligned}
\tilde{\mathbf{e}}_s = G_{enc}({\rm Emb}(x_s),\tilde{\mathbf{e}}_{s-1})
\end{aligned}
\end{equation}
where $\tilde{\mathbf{e}}_s$ is the hidden state, which represents the word $x_s$ from the perspective of the MC model. Instead of the conventional seq2seq models that directly send the results of the equation (\ref{eq4}) to the decoder , we concatenate $\tilde{\mathbf{e}}_s$ and $\tilde{\mathbf{h}}_s$ and feed them into an integration layer:
\begin{equation}
\begin{aligned}
\bar{\mathbf{h}}_s = F_{int}([\tilde{\mathbf{h}}_s;\tilde{\mathbf{e}}_s],\bar{\mathbf{h}}_{s-1})
\end{aligned}
\end{equation}
where $F_{int}$ is a uni-directional LSTM, $[;]$ means concatenation. $\{\bar{\mathbf{h}}_s\}_{s=1}^{T_x}$ are the contextual representations of the sentence $x$ which contain the information of the machine comprehension knowledge as well.

\subsection{Decoder \& Attention Mechanism}
Initialized by the representations obtained from the encoder, the decoder with an attention mechanism receives the word embedding of the previous word (while training, it is the previous word of the reference sentence; while testing, it is the previous generated word) at each step and generates next word. The decoder states are computed via:
\begin{equation}
\begin{aligned}
\bar{\mathbf{h}}_t = F_{dec}({\rm Emb}(y_{t-1}),\bar{\mathbf{h}}_{t-1})
\end{aligned}
\end{equation}
where $F_{dec}$ is a unidirectional LSTM, $y_t$ is the $t$-th generated word, $\bar{\mathbf{h}}_s$ is the hidden state. For most seq2seq attentional models, the attention steps can be summarized by the equations below:
\begin{equation}
\begin{aligned}
\alpha_{ts} = \frac{{\rm exp(score}(\bar{\mathbf{h}}_s,\bar{\mathbf{h}}_t))}{\sum_{s' = 1}^{T_x}{\rm exp(score}(\bar{\mathbf{h}}_{s'},\bar{\mathbf{h}}_t))}
\end{aligned}
\end{equation}
\begin{equation}
\begin{aligned}
\mathbf{c}_t = \sum_{s} \alpha_{ts} \bar{\mathbf{h}}_s
\end{aligned}
\end{equation}
\begin{equation}
\begin{aligned}
\mathbf{a}_t = g_a (\mathbf{c}_t, \bar{\mathbf{h}}_t) = {\rm tanh}(\mathbf{W}_a[\mathbf{c}_t; \bar{\mathbf{h}}_t] + \mathbf{b}_a)
\end{aligned}
\end{equation}
Here, $\mathbf{c}_t$ is the source-side context vector, the attention vector $\mathbf{a}_t$ is used to derive the softmax logit and loss, $\mathbf{W}_a$ and $\mathbf{b}_a$ are trainable parameters, the function $g_a$ can also take other forms. ${\rm score}$ is referred as a \emph{content-based} function, usually implemented as a feed forward network with one hidden layer.

For the common seq2seq models, the attention vector $\mathbf{a}_t$ is then fed through the softmax layer to produce the predictive distribution formulated as:
\begin{equation}
\label{eq13}
\begin{aligned}
P(y_t|y_{<t}, x) \propto {\rm softmax}(\mathbf{W}_p \mathbf{a}_t + \mathbf{b}_p)
\end{aligned}
\end{equation}
In our MacNet, however, we additionally send the attention vector $\mathbf{a}_t$ into the modeling layer of the pre-trained MC model in the equation (\ref{eq3}) to deeply capture the interaction of the source and the target states:
\begin{equation}
\begin{aligned}
\mathbf{r}_t = G_{model} (\mathbf{a}_t,\mathbf{r}_{t-1})
\end{aligned}
\end{equation}
where $\mathbf{r}_t$ is another attention state with the augmentation of machine comprehension knowledge. We combine the results of the two attention vectors and the equation (\ref{eq13}) becomes:
\begin{equation}
\label{eq14}
\begin{aligned}
P(y_t|y_{<t}, x) \propto {\rm softmax}(\mathbf{W}_p \mathbf{a}_t + \mathbf{W}_q \mathbf{r}_t + \mathbf{b}_p)
\end{aligned}
\end{equation}
where $\mathbf{W}_p, \mathbf{W}_q$ and $\mathbf{b}_p$ are all trainable parameters. The modeling layer helps deeply understand the interaction of the contextual information of the output sequence, which is different from the encoding layer whose inputs are independent source sentences.

\subsection{Training}
Denote $\Theta$ as all the parameters to be learned in the framework, $D$ as the training dataset that contains source-target sequence pairs. The training process aims at seeking the optimal paramaters $\Theta^{*}$ that encodes the source sequence and provides an output sentence as close as the target sentence. For the formula form, the most popular objective is the maximum log likelihood estimation~\citep{bahdanau2015neural, xia2017deliberation}:
\begin{equation}
\label{eq15}
\begin{aligned}
\Theta^{*} & = \mathop{\arg\max}_{\Theta} \sum_{(x,y) \in D} P(y|x; \Theta)\\
& = \mathop{\arg\max}_{\Theta} \sum_{(x,y) \in D} \sum_{t = 1}^{T_y} {\rm log} P(y_t|y_{<t}, x; \Theta)
\end{aligned}
\end{equation}

However, this results in the high frequency of some commonly used expressions such as ``I don't know" in the output sentences because of the nature of the class imbalance in the corpus. Inspired by the \emph{focal loss}~\citep{lin2017focal}, which is recently proposed to solve the foreground-background class imbalance in the task of object detection, we add a modulating factor to the above cross entropy loss. Simplifying $P(y_t|y_{<t}, x; \Theta)$ as $p_t$, we modify the equation (\ref{eq15}) as:
\begin{equation}
\begin{aligned}
\Theta^{*} & =  \mathop{\arg\max}_{\Theta} \sum_{(x,y) \in D} \sum_{t = 1}^{T_y} (1- p_t)^{\gamma}{\rm log} (p_t)
\end{aligned}
\end{equation}
where $\gamma$ is a tunable focusing parameter. In this case, the focusing parameter smoothly adjusts the rate at which high-frequency phrases are down-weighted.

\section{Experiments}
\subsection{Machine Comprehension}
We use the Stanford Question Answering Dataset (SQuAD)\citep{rajpurkar2016squad} as our training set\footnote{The SQuAD dataset is referred at: \url{https://rajpurkar.github.io/SQuAD-explorer/}}, which has 100,000+ questions posed by crowd workers on 536 Wikipedia articles. The hidden state size of the LSTM is set as 100, and we select the 300d Glove as the word embeddings. We use 100 one dimensional filters for CNN in the character level embedding, with width of 5 for each one. The dropout ratio is 0.2. We use the AdaDelta~\citep{zeiler2012adadelta} optimizer with an initial learning rate as 0.001. Our MC model achieves 67.08 of Exact Match (EM) and 76.79 of F1 score on the SQuAD development dataset.

\subsection{Application to Neural Machine Translation}
We first evaluate our method on the neural machine translation (NMT) task, which requires to encode a source language sentence and predict a target language sentence. We use the architecture from \citep{luong2015effective} as our baseline framework with the GNMT~\citep{45610} attention to parallelize the decoder's computation. The datasets for our evaluation are the WMT translation tasks between English and German in both directions. Translation performances are reported in case-sensitive BLEU~\citep{papineni2002bleu} on newstest2014\footnote{ \url{http://www.statmt.org/wmt14/translation-task.html}} and newstest2015\footnote{ \url{http://www.statmt.org/wmt15/translation-task.html}}. 

\paragraph{Implementation details: }When training our NMT systems, we split the data into subword units using BPE~\citep{sennrich2016neural}. We train 4-layer LSTMs of 1024 units with bidirectional encoder, embedding dimension is 1024. We use a fully connected layer to transform the input vector size for the transferred neural networks. The model is trained with stochastic gradient descent with a learning rate that began at 1. We train for 340K steps; after 170K steps, we start halving learning rate every 17K step. Our batch size is set as 128, the dropout rate is 0.2. For the focal loss, the $\gamma$ is set to be 5.

\begin{table*}[t]
	\begin{center}
		\begin{tabular}{lccccp{1cm}cp{1cm}cp{1cm}cp{1cm}}
			\toprule
			\multicolumn{1}{c}{\multirow{2}{*}{\textbf{NMT Systems}}} & \multicolumn{2}{c}{\textbf{WMT14}} &  \multicolumn{2}{c}{\textbf{WMT15}} \\ \cline{2-5} 
			\multicolumn{1}{c}{}  & \textbf{En}$\to$\textbf{De} & 	\textbf{De$\to$En} & \textbf{En}$\to$\textbf{De} & 	\textbf{De$\to$En} \\
			\midrule
			Baseline  & 22.1 & 26.0 & 24.5 & 27.5\\ \hline
			Baseline + Encoding Layer &  23.2 & 27.0 &  25.3& 28.3\\ 
			Baseline + Modeling Layer & 22.4 & 26.4 & 24.8 & 27.8\\
			Baseline + Encoding Layer + Modeling Layer  & 23.4 & 27.3 & 25.6 & 28.5 \\
			Baseline + Random Initalized Framework & 21.6 & 25.6 & 24.2 & 27.0\\  \hline
			\textbf{Baseline + MacNet} & \textbf{24.2} & \textbf{28.1} & \textbf{26.3} & \textbf{29.4} \\
			\bottomrule
		\end{tabular}
		\vspace{2mm}
	\end{center}
	\caption{\label{tab2} BLEU scores on official test sets (WMT English-German for \texttt{newstest2014} and \texttt{newstest2015}). In the top part, we show the performance of our baseline model; In the medium part, we present the ablation experiments; In the bottom part, we show the effectiveness of our MacNet.}       
\end{table*}

\begin{wraptable}{r}{0.5\textwidth}
	\begin{tabular}{lcc}
		\toprule
		\textbf{MC Attention}	& \textbf{EM} & \textbf{BLEU} \\
		\midrule
		Context to Query Attention & 63.3 & 25.1\\ 
		Query to Context Attention & 56.9 & 25.3\\ 
		BiDAF & 67.1 & 27.5 \\
		BiDAF + Self-Attention & 68.2 & 27.4\\ 
		BiDAF + Memory Network & 68.5 & 27.6\\ 
		\bottomrule
	\end{tabular}
	\vspace{2mm}
	\caption{\label{tab3}Performance with different pre-trained machine comprehension models for our NMT model on De$\to$En of WMT'14. \textbf{EM} means the exact match score, which represents the performance of the MC model on the SQuAD dev set, \textbf{BLEU} is the results of our NMT model.}     
\end{wraptable}

\paragraph{Results: }As shown in the Table \ref{tab2}, the baseline NMT model on all of the datasets performs much better with the help of our MacNet framework. In the medium part, we conduct an ablation experiment to evaluate the individual contribution of each component of our model. Both of the encoding layer and the modeling layer demonstrates their effectiveness when we ablate other modules. When we add both of them (still without the focal loss), the BLEU scores on all the test sets rise at least 1 point, which shows the significance of the transferred knowledge. Finally, we add the architecture of the encoding layer and the modeling layer to the baseline model but initialize them randomly as its other RNN layers. We observe that the performance drops around 0.5\%, which indicates that the machine comprehension knowledge has deep connections with the machine translation tasks. From the ablation experiments we found that the improvement of the modeling layer in our architecture is a bit modest, but we believe transferring high-level networks (\emph{e.g.} the modeling layer) can help a lot with a more suitable structure because those networks contains deeper semantic
knowledge and more abstractive information compared with the lower-level layers (\emph{e.g.} encoding layer).

\begin{wrapfigure}{r}{0.45\textwidth}
	
	\includegraphics[width=0.48 \textwidth]{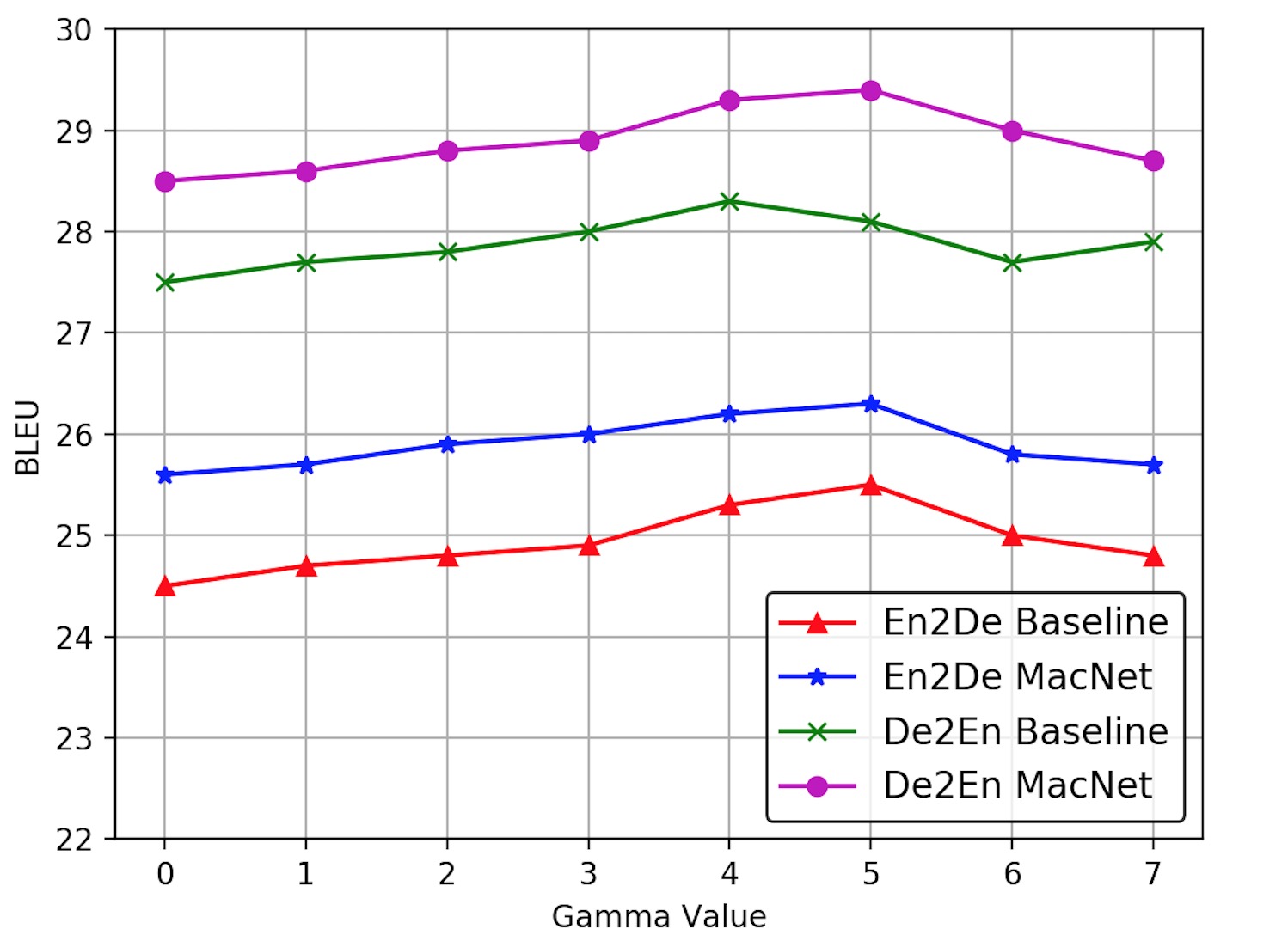}
	\caption{\label{fig2}Performance on the WMT'15 with different $\gamma$ values.}
\end{wrapfigure}

In the Table \ref{tab3}, we explore how different choices of the attention architectures ($f_{att}$ in the equation (\ref{eq2}), which is usually the discrimination of different MC models) of the MC models impact the performance of our method. We first follow \citep{seo2017bidirectional} to separate the two directions of the attention in BiDAF and use them to take place of the original attention mechanism respectively. Their performance on the machine comprehension task drops a lot, and it seems to affect the results of the NMT models as well. We then add the self-attention, which is proposed to fuse the context into itself, is widely used by many MC methods~\citep{rnet,weissenborn2017making}. Unfortunately, the result of the NMT model fails to keep pace with the performance of its pre-train MC model. Finally, we apply memory network, which is also very popular among MC models~\citep{pan2017memen,hu2017reinforced}, the performance on the SQuAD rises a lot but the NMT result is similar to the original model. This series of experiments denote that the model's performance with our MacNet is not always in positive correlation to the improvement of the MC architecture. We conjecture that it might depend on many potential factors such as the complexity of the extracted parts, the heterogeneity of different tasks, \emph{etc}.

In the Figure \ref{fig2}, we present the models on the WMT'15 with different $\gamma$ to show how the focal loss affects the performance. As we can see, the models increase as the $\gamma$ enlarges until it arrives 4 or 5. Afterwards, the performance gets worse when we raise the $\gamma$, which means the modulating factor is close to zero so that its benefit is limited.

\begin{table*}[t]
	\begin{center}
		\begin{tabular}{lccccccp{0.8cm}cp{0.8cm}cp{0.8cm}cp{0.8cm}cp{0.8cm}cp{0.8cm}}
			\toprule
			\multicolumn{1}{c}{\multirow{2}{*}{\textbf{Summarization Models}}} & \multicolumn{3}{c}{\textbf{CNN/Daily Mail}} &  \multicolumn{3}{c}{\textbf{Gigaword}} \\ \cline{2-7} 
			\multicolumn{1}{c}{}  & \textbf{RG-1} & 	\textbf{RG-2} & \textbf{RG-L} & 	\textbf{RG-1} & \textbf{RG-2} & \textbf{RG-L} \\
			\midrule
			words-lvt5k\citep{nallapati2016abstractive}  & 35.46$^\dag$ & 13.30$^\dag$ & 32.65$^\dag$ & 35.30$^\dag$ & 16.64$^\dag$ & 32.62$^\dag$\\
			SummaRuNNer\citep{nallapati2017summarunner} & 39.60$^\dag$ & 16.20$^\dag$ & 35.30$^\dag$ & -- & -- & -- \\ 	
			ConvS2S\citep{gehring2017convolutional} & -- & -- & -- & 35.88$^\dag$ & 17.48$^\dag$ & 33.29$^\dag$\\
			SEASS\citep{zhou2017selective}  &  -- & -- & -- & 36.15$^\dag$ & 17.54$^\dag$ & 33.63$^\dag$ \\
			RL with intra-attn\citep{paulus2017deep} & \textbf{41.16}$^\dag$ & 15.75$^\dag$ & \textbf{39.08}$^\dag$ &  -- & -- & --\\ \hline 		
			Pointer-Generator\citep{see2017get} & 39.69$^{~}$ & 17.26$^{~}$ & 36.38$^{~}$ & 36.44$^{~}$ & 17.26$^{~}$ & 33.92$^{~}$\\   
			Pointer-Generator + Encoding Layer & 40.38$^{~}$ & 17.75$^{~}$ & 37.24$^{~}$ & 37.30$^{~}$ & 17.83$^{~}$ & 34.41$^{~}$\\ 
			Pointer-Generator + Modeing Layer & 39.92$^{~}$ & 17.58$^{~}$ & 36.65$^{~}$ & 36.85$^{~}$ & 17.45$^{~}$ & 34.12$^{~}$\\ 
			\textbf{Pointer-Generator + MacNet} & 40.87$^{~}$ & \textbf{18.02}$^{~}$ & 37.54$^{~}$ & \textbf{37.97}$^{~}$ & \textbf{18.16}$^{~}$ & \textbf{34.93}$^{~}$\\
			\bottomrule
		\end{tabular}
		\vspace{2mm}
	\end{center}
	\caption{\label{tab4} ROUGE F$_1$ evaluation results on the CNN/Daily Mail test set and the English Gigaword test set. RG in the Table denotes ROUGE. Results with $^\dag$ mark are taken from the corresponding papers. The bottom part of the Table shows the performance of our MacNet and the ablation results.}       
\end{table*}

\subsection{Application to Text Summarization}
We then verify the effectiveness of our MacNet on the abstractive text summarization, which is also a typical application of the sequence-to-sequence model. We use the Pointer-Generator Networks\citep{see2017get} as our baseline model, which applies the encoder-decoder architecture and is one of the state-of-the-art models for the text summarization. The evaluation metric is reported with the F$_{1}$ scores for ROUGE-1, ROUGE-2 and ROUGE-L~\citep{lin2004rouge}. We evaluate our method on two high-quality datasets, \emph{CNN/Daily Mail}~\citep{hermann2015teaching} and \emph{Gigaword}~\citep{rush2015neural}. For the CNN/Daily Mail dataset, we use scripts\footnote{\url{https://github.com/abisee/cnn-dailymail}} supplied by \citet{see2017get} to pre-process the data, which contains 287k training pairs, 13k validation pairs and 11k test pairs. For the English Gigaword dataset, we use the script\footnote{\url{https://github.com/facebook/NAMAS}} released by \citet{rush2015neural} to pre-process and obtain 3.8M training pairs, 189k development set for testing.

\paragraph{Implementation details: } Our training hyperparameters are similar to the Pointer-Generator Networks experiments, while some important details are as follows. The input and output vocabulary size is 50k, the hidden state size is 256. The word embedding size is 128, and we use a fully connected layer to transform the input vector size for the transferred neural networks. We train using Adagrad~\citep{duchi2011adaptive} with learning rate 0.15 and an initial accumulator value of 0.1. The $\gamma$ is set as 3.

\paragraph{Results: }Table \ref{tab4} shows the performance of our methods and the competing approaches on both datasets. Compared to the original Pointer-Generator model, the results with our MacNet architecture outperform around 0.7\% $\sim$ 1.5\% on all kinds of the ROUGE scores. Especially, our approach achieves the state-of-the-art results on all the metrics on Gigaword and the ROUGE-2 on CNN/Daily Mail dataset. Similar to the NMT task, the encoding layer contributes most of the improvement, while the modeling layer also has stable gains in each evaluations. 

In the Table \ref{tab5}, we present some summaries produced by our model and the original Pointer-Generator model. In the first example, the summary given by the Pointer-Generator model doesn't make sense from the perspective of logic, while our model accurately summarizes the article and even provides with more details. In the second example, although the original PG model produces a logical sentence, the output sentence expresses completely different meanings from the information in the article. Our method, however, correctly comprehends the article and provides with a high-quality summary sentence.

\begin{table*}[t]
	\begin{center}
		\begin{tabular}{l}
			\toprule
			\textbf{Article: }{\small Israeli warplanes raided Hezbollah targets in south Lebanon after guerrillas killed two militiamen}\\ {\small and wounded seven other troops on Wednesday,
				police said.}\\
			\textbf{Reference: }{\small Israeli warplanes raid south Lebanon.}\\ 
			\textbf{PG + MacNet: }{\small Israeli warplanes attack Hezbollah targets in south Lebanon.}\\ 
			\textbf{PG: }{\small Hezbollah targets Hezbollah targets in south Lebanon.} \\ \hline
			\textbf{Article: }{\small The dollar racked up some clear gains on Wednesday on the London forex market as operators}\\ {\small waited for the outcome of talks between the White House and Congress on raising the national debt ceiling}\\ {\small and on cutting the American budget deficit.}\\
			\textbf{Reference: }{\small Dollar gains as market eyes US debt and budget talks.}\\
			\textbf{PG + MacNet: }: {\small Dollar racked up some clear gains.}\\
			\textbf{PG}: {\small London forex market racked gains.}\\
			\bottomrule
		\end{tabular}
	\end{center}
	\caption{\label{tab5} Examples of summaries on English Gigaword, \textbf{PG} denotes the Pointer-Generator model.}       
\end{table*}

\section{Conclusion}
In this paper, we propose \emph{MacNet}, which is a supplementary framework for the sequence-to-sequence tasks. We transfer the knowledge from the machine comprehension task to a variety of seq2seq tasks to augment the text understanding of the models. The experimental evaluation shows that our method significantly improves the performance of the baseline models on several benchmark datasets for different NLP tasks. We hope this work can encourage further research into the transfer learning of multi-layer neural networks, and the future works involve the choice of other transfer learning sources and the transfer learning between different domains such as NLP, CV, \emph{etc}.

\subsubsection*{Acknowledgments}
This work was supported in part by the National Nature Science Foundation of China (Grant Nos: 61751307, 61602405 and U1611461) and in part by the National Youth Top-notch Talent Support Program. The experiments are supported by Chengwei Yao in the Experiment Center of the College of Computer Science and Technology, Zhejiang University.

\small

\bibliography{macnet_neurips}

\begin{thebibliography}{}

\bibitem[\protect\citeauthoryear{Bahdanau \bgroup et al\mbox.\egroup
  }{2015}]{bahdanau2015neural}
Bahdanau, D.; Cho, K.; Bengio, Y.; et~al.
\newblock 2015.
\newblock Neural machine translation by jointly learning to align and
  translate.
\newblock In {\em ICLR}.

\bibitem[\protect\citeauthoryear{Chan \bgroup et al\mbox.\egroup
  }{2016}]{chan2016listen}
Chan, W.; Jaitly, N.; Le, Q.; and Vinyals, O.
\newblock 2016.
\newblock Listen, attend and spell: A neural network for large vocabulary
  conversational speech recognition.
\newblock In {\em Acoustics, Speech and Signal Processing (ICASSP), 2016 IEEE
  International Conference on},  4960--4964.

\bibitem[\protect\citeauthoryear{Cho \bgroup et al\mbox.\egroup
  }{2014}]{cho2014learning}
Cho, K.; van Merrienboer, B.; Gulcehre, C.; Bahdanau, D.; Bougares, F.;
  Schwenk, H.; and Bengio, Y.
\newblock 2014.
\newblock Learning phrase representations using rnn encoder--decoder for
  statistical machine translation.
\newblock In {\em EMNLP},  1724--1734.

\bibitem[\protect\citeauthoryear{Collobert \bgroup et al\mbox.\egroup
  }{2011}]{collobert2011natural}
Collobert, R.; Weston, J.; Bottou, L.; Karlen, M.; Kavukcuoglu, K.; and Kuksa,
  P.
\newblock 2011.
\newblock Natural language processing (almost) from scratch.
\newblock {\em Journal of Machine Learning Research} 12(Aug):2493--2537.

\bibitem[\protect\citeauthoryear{Cui \bgroup et al\mbox.\egroup
  }{2017}]{cui2017attention}
Cui, Y.; Chen, Z.; Wei, S.; Wang, S.; Liu, T.; and Hu, G.
\newblock 2017.
\newblock Attention-over-attention neural networks for reading comprehension.
\newblock In {\em ACL}, volume~1,  593--602.

\bibitem[\protect\citeauthoryear{Duchi \bgroup et al\mbox.\egroup
  }{2011}]{duchi2011adaptive}
Duchi, J.; Hazan, E.; Singer, Y.; et~al.
\newblock 2011.
\newblock Adaptive subgradient methods for online learning and stochastic
  optimization.
\newblock {\em Journal of Machine Learning Research} 12(Jul):2121--2159.

\bibitem[\protect\citeauthoryear{Gehring \bgroup et al\mbox.\egroup
  }{2017}]{gehring2017convolutional}
Gehring, J.; Auli, M.; Grangier, D.; Yarats, D.; and Dauphin, Y.~N.
\newblock 2017.
\newblock Convolutional sequence to sequence learning.
\newblock {\em arXiv preprint arXiv:1705.03122}.

\bibitem[\protect\citeauthoryear{Glorot \bgroup et al\mbox.\egroup
  }{2011}]{glorot2011domain}
Glorot, X.; Bordes, A.; Bengio, Y.; et~al.
\newblock 2011.
\newblock Domain adaptation for large-scale sentiment classification: A deep
  learning approach.
\newblock In {\em ICML},  513--520.

\bibitem[\protect\citeauthoryear{Guo, Pasunuru, and Bansal}{2018}]{guo2018soft}
Guo, H.; Pasunuru, R.; and Bansal, M.
\newblock 2018.
\newblock Soft layer-specific multi-task summarization with entailment and
  question generation.
\newblock {\em arXiv preprint arXiv:1805.11004}.

\bibitem[\protect\citeauthoryear{Hermann \bgroup et al\mbox.\egroup
  }{2015}]{hermann2015teaching}
Hermann, K.~M.; Kocisky, T.; Grefenstette, E.; Espeholt, L.; Kay, W.; Suleyman,
  M.; and Blunsom, P.
\newblock 2015.
\newblock Teaching machines to read and comprehend.
\newblock In {\em NIPS},  1693--1701.

\bibitem[\protect\citeauthoryear{Hu \bgroup et al\mbox.\egroup
  }{2017}]{hu2017reinforced}
Hu, M.; Peng, Y.; Qiu, X.; et~al.
\newblock 2017.
\newblock Reinforced mnemonic reader for machine comprehension.
\newblock {\em CoRR, abs/1705.02798}.

\bibitem[\protect\citeauthoryear{Joshi \bgroup et al\mbox.\egroup
  }{2017}]{joshi2017triviaqa}
Joshi, M.; Choi, E.; Weld, D.; and Zettlemoyer, L.
\newblock 2017.
\newblock Triviaqa: A large scale distantly supervised challenge dataset for
  reading comprehension.
\newblock In {\em ACL}.

\bibitem[\protect\citeauthoryear{Kim}{2014}]{kim2014convolutional}
Kim, Y.
\newblock 2014.
\newblock Convolutional neural networks for sentence classification.
\newblock In {\em EMNLP},  1746--1751.

\bibitem[\protect\citeauthoryear{Li \bgroup et al\mbox.\egroup
  }{2017}]{li2017modeling}
Li, J.; Xiong, D.; Tu, Z.; Zhu, M.; Zhang, M.; and Zhou, G.
\newblock 2017.
\newblock Modeling source syntax for neural machine translation.
\newblock In {\em ACL}, volume~1,  688--697.

\bibitem[\protect\citeauthoryear{Lin \bgroup et al\mbox.\egroup
  }{2017}]{lin2017focal}
Lin, T.-Y.; Goyal, P.; Girshick, R.; He, K.; and Dollar, P.
\newblock 2017.
\newblock Focal loss for dense object detection.
\newblock In {\em ICCV},  2980--2988.

\bibitem[\protect\citeauthoryear{Lin}{2004}]{lin2004rouge}
Lin, C.-Y.
\newblock 2004.
\newblock Rouge: A package for automatic evaluation of summaries.
\newblock {\em Text Summarization Branches Out}.

\bibitem[\protect\citeauthoryear{Luong \bgroup et al\mbox.\egroup
  }{2015}]{luong2015effective}
Luong, T.; Pham, H.; Manning, C.~D.; et~al.
\newblock 2015.
\newblock Effective approaches to attention-based neural machine translation.
\newblock In {\em EMNLP},  1412--1421.

\bibitem[\protect\citeauthoryear{McCann \bgroup et al\mbox.\egroup
  }{2017}]{mccann2017learned}
McCann, B.; Bradbury, J.; Xiong, C.; and Socher, R.
\newblock 2017.
\newblock Learned in translation: Contextualized word vectors.
\newblock In {\em NIPS},  6297--6308.

\bibitem[\protect\citeauthoryear{Min, Seo, and
  Hajishirzi}{2017}]{min2017question}
Min, S.; Seo, M.; and Hajishirzi, H.
\newblock 2017.
\newblock Question answering through transfer learning from large fine-grained
  supervision data.
\newblock In {\em ACL}.

\bibitem[\protect\citeauthoryear{Nallapati \bgroup et al\mbox.\egroup
  }{2016}]{nallapati2016abstractive}
Nallapati, R.; Zhou, B.; dos Santos, C.; Gulcehre, C.; and Xiang, B.
\newblock 2016.
\newblock Abstractive text summarization using sequence-to-sequence rnns and
  beyond.
\newblock In {\em Proceedings of The 20th SIGNLL Conference on Computational
  Natural Language Learning},  280--290.

\bibitem[\protect\citeauthoryear{Nallapati \bgroup et al\mbox.\egroup
  }{2017}]{nallapati2017summarunner}
Nallapati, R.; Zhai, F.; Zhou, B.; et~al.
\newblock 2017.
\newblock Summarunner: A recurrent neural network based sequence model for
  extractive summarization of documents.
\newblock In {\em AAAI},  3075--3081.

\bibitem[\protect\citeauthoryear{Pan \bgroup et al\mbox.\egroup
  }{2017}]{pan2017memen}
Pan, B.; Li, H.; Zhao, Z.; Cao, B.; Cai, D.; and He, X.
\newblock 2017.
\newblock Memen: Multi-layer embedding with memory networks for machine
  comprehension.
\newblock {\em arXiv preprint arXiv:1707.09098}.

\bibitem[\protect\citeauthoryear{Pan \bgroup et al\mbox.\egroup
  }{2018}]{pan2018discourse}
Pan, B.; Yang, Y.; Zhao, Z.; Zhuang, Y.; Cai, D.; and He, X.
\newblock 2018.
\newblock Discourse marker augmented network with reinforcement learning for
  natural language inference.
\newblock In {\em Proceedings of the 56th Annual Meeting of the Association for
  Computational Linguistics (Volume 1: Long Papers)}, volume~1,  989--999.

\bibitem[\protect\citeauthoryear{Papineni \bgroup et al\mbox.\egroup
  }{2002}]{papineni2002bleu}
Papineni, K.; Roukos, S.; Ward, T.; and Zhu, W.-J.
\newblock 2002.
\newblock Bleu: a method for automatic evaluation of machine translation.
\newblock In {\em ACL},  311--318.
\newblock Association for Computational Linguistics.

\bibitem[\protect\citeauthoryear{Paulus \bgroup et al\mbox.\egroup
  }{2017}]{paulus2017deep}
Paulus, R.; ~, C.; Socher, R.; et~al.
\newblock 2017.
\newblock A deep reinforced model for abstractive summarization.
\newblock {\em arXiv preprint arXiv:1705.04304}.

\bibitem[\protect\citeauthoryear{Pennington \bgroup et al\mbox.\egroup
  }{2014}]{pennington2014glove}
Pennington, J.; Socher, R.; Manning, C.~D.; et~al.
\newblock 2014.
\newblock Glove: Global vectors for word representation.
\newblock In {\em EMNLP},  1532--1543.

\bibitem[\protect\citeauthoryear{Rajpurkar \bgroup et al\mbox.\egroup
  }{2016}]{rajpurkar2016squad}
Rajpurkar, P.; Zhang, J.; Lopyrev, K.; and Liang, P.
\newblock 2016.
\newblock Squad: 100,000+ questions for machine comprehension of text.
\newblock In {\em EMNLP},  2383–2392.

\bibitem[\protect\citeauthoryear{Rush \bgroup et al\mbox.\egroup
  }{2015}]{rush2015neural}
Rush, A.~M.; Chopra, S.; Weston, J.; et~al.
\newblock 2015.
\newblock A neural attention model for abstractive sentence summarization.
\newblock In {\em EMNLP},  379--389.

\bibitem[\protect\citeauthoryear{See \bgroup et al\mbox.\egroup
  }{2017}]{see2017get}
See, A.; Liu, P.~J.; Manning, C.~D.; et~al.
\newblock 2017.
\newblock Get to the point: Summarization with pointer-generator networks.
\newblock In {\em ACL}, volume~1,  1073--1083.

\bibitem[\protect\citeauthoryear{Sennrich \bgroup et al\mbox.\egroup
  }{2016}]{sennrich2016neural}
Sennrich, R.; Haddow, B.; Birch, A.; et~al.
\newblock 2016.
\newblock Neural machine translation of rare words with subword units.
\newblock In {\em ACL}.

\bibitem[\protect\citeauthoryear{Seo \bgroup et al\mbox.\egroup
  }{2017}]{seo2017bidirectional}
Seo, M.; Kembhavi, A.; Farhadi, A.; and Hajishirzi, H.
\newblock 2017.
\newblock Bidirectional attention flow for machine comprehension.
\newblock In {\em ICLR}.

\bibitem[\protect\citeauthoryear{Shen \bgroup et al\mbox.\egroup
  }{2017}]{shen2017reasonet}
Shen, Y.; Huang, P.-S.; Gao, J.; and Chen, W.
\newblock 2017.
\newblock Reasonet: Learning to stop reading in machine comprehension.
\newblock In {\em KDD},  1047--1055.
\newblock ACM.

\bibitem[\protect\citeauthoryear{Shi \bgroup et al\mbox.\egroup
  }{2016}]{shi2016does}
Shi, X.; Padhi, I.; Knight, K.; et~al.
\newblock 2016.
\newblock Does string-based neural mt learn source syntax?
\newblock In {\em ACL},  1526--1534.

\bibitem[\protect\citeauthoryear{Sutskever \bgroup et al\mbox.\egroup
  }{2014}]{sutskever2014sequence}
Sutskever, I.; Vinyals, O.; Le, Q.~V.; et~al.
\newblock 2014.
\newblock Sequence to sequence learning with neural networks.
\newblock In {\em NIPS},  3104--3112.

\bibitem[\protect\citeauthoryear{Vinyals \bgroup et al\mbox.\egroup
  }{2015}]{vinyals2015pointer}
Vinyals, O.; Fortunato, M.; Jaitly, N.; et~al.
\newblock 2015.
\newblock Pointer networks.
\newblock In {\em NIPS},  2692--2700.

\bibitem[\protect\citeauthoryear{Wang \bgroup et al\mbox.\egroup
  }{2016}]{wang2016multi}
Wang, Z.; Mi, H.; Hamza, W.; and Florian, R.
\newblock 2016.
\newblock Multi-perspective context matching for machine comprehension.
\newblock {\em arXiv preprint arXiv:1612.04211}.

\bibitem[\protect\citeauthoryear{Wang \bgroup et al\mbox.\egroup }{2017}]{rnet}
Wang, W.; Yang, N.; Wei, F.; Chang, B.; and Zhou, M.
\newblock 2017.
\newblock Gated self-matching networks for reading comprehension and question
  answering.
\newblock In {\em ACL}.

\bibitem[\protect\citeauthoryear{Weissenborn \bgroup et al\mbox.\egroup
  }{2017}]{weissenborn2017making}
Weissenborn, D.; Wiese, G.; Seiffe, L.; et~al.
\newblock 2017.
\newblock Making neural qa as simple as possible but not simpler.
\newblock In {\em CoNLL},  271--280.

\bibitem[\protect\citeauthoryear{Williams \bgroup et al\mbox.\egroup
  }{2017}]{williams2017hybrid}
Williams, J.~D.; Asadi, K.; Zweig, G.; et~al.
\newblock 2017.
\newblock Hybrid code networks: practical and efficient end-to-end dialog
  control with supervised and reinforcement learning.
\newblock In {\em ACL}, volume~1,  665--677.

\bibitem[\protect\citeauthoryear{Wu \bgroup et al\mbox.\egroup }{2016}]{45610}
Wu, Y.; Schuster, M.; Chen, Z.; Le, Q.~V.; Norouzi, M.; Macherey, W.; Krikun,
  M.; Cao, Y.; Gao, Q.; Macherey, K.; Klingner, J.; Shah, A.; Johnson, M.; Liu,
  X.; Łukasz Kaiser; Gouws, S.; Kato, Y.; Kudo, T.; Kazawa, H.; Stevens, K.;
  Kurian, G.; Patil, N.; Wang, W.; Young, C.; Smith, J.; Riesa, J.; Rudnick,
  A.; Vinyals, O.; Corrado, G.; Hughes, M.; and Dean, J.
\newblock 2016.
\newblock Google's neural machine translation system: Bridging the gap between
  human and machine translation.
\newblock {\em CoRR} abs/1609.08144.

\bibitem[\protect\citeauthoryear{Xia \bgroup et al\mbox.\egroup
  }{2017}]{xia2017deliberation}
Xia, Y.; Tian, F.; Wu, L.; Lin, J.; Qin, T.; Yu, N.; and Liu, T.-Y.
\newblock 2017.
\newblock Deliberation networks: Sequence generation beyond one-pass decoding.
\newblock In {\em NIPS},  1782--1792.

\bibitem[\protect\citeauthoryear{Xiong \bgroup et al\mbox.\egroup
  }{2017}]{xiong2017dynamic}
Xiong, C.; Zhong, V.; Socher, R.; et~al.
\newblock 2017.
\newblock Dynamic coattention networks for question answering.
\newblock {\em ICLR 2017}.

\bibitem[\protect\citeauthoryear{Xiong \bgroup et al\mbox.\egroup
  }{2018}]{xiong2018dcn}
Xiong, C.; Zhong, V.; Socher, R.; et~al.
\newblock 2018.
\newblock {DCN}+: Mixed objective and deep residual coattention for question
  answering.
\newblock In {\em ICLR}.

\bibitem[\protect\citeauthoryear{Xu \bgroup et al\mbox.\egroup
  }{2015}]{xu2015show}
Xu, K.; Ba, J.; Kiros, R.; Cho, K.; Courville, A.; Salakhudinov, R.; Zemel, R.;
  and Bengio, Y.
\newblock 2015.
\newblock Show, attend and tell: Neural image caption generation with visual
  attention.
\newblock In {\em International Conference on Machine Learning},  2048--2057.

\bibitem[\protect\citeauthoryear{Zeiler}{2012}]{zeiler2012adadelta}
Zeiler, M.~D.
\newblock 2012.
\newblock Adadelta: an adaptive learning rate method.
\newblock {\em arXiv preprint arXiv:1212.5701}.

\bibitem[\protect\citeauthoryear{Zhou \bgroup et al\mbox.\egroup
  }{2017}]{zhou2017selective}
Zhou, Q.; Yang, N.; Wei, F.; and Zhou, M.
\newblock 2017.
\newblock Selective encoding for abstractive sentence summarization.
\newblock In {\em ACL},  1095--1104.

\end{thebibliography}
\bibliographystyle{neurips_2018}

\end{document}